# The Arabic AI Fingerprint: Stylometric Analysis and Detection of Large Language Models Text

Maged S. Al-Shaibani, Moataz Ahmed[*]

[a]*SDAIA-KFUPM Joint Research Center for Artificial Intelligence, King Fahd University of Petroleum and Minerals, Dhahran, 31261, Eastern Province, Saudi Arabia*

**Abstract**

Large Language Models (LLMs) have achieved unprecedented capabilities in generating human-like text, posing subtle yet significant challenges for information integrity across critical domains, including education, social media, and academia, enabling sophisticated misinformation campaigns, compromising healthcare guidance, and facilitating targeted propaganda. This challenge becomes severe, particularly in under-explored and low-resource languages like Arabic. This paper presents a comprehensive investigation of Arabic machine-generated text, examining multiple generation strategies (generation from the title only, content-aware generation, and text refinement) across diverse model architectures (ALLaM, Jais, Llama, and GPT-4) in academic, and social media domains. Our stylometric analysis reveals distinctive linguistic patterns differentiating human-written from machine-generated Arabic text across these varied contexts. Despite their human-like qualities, we demonstrate that LLMs produce detectable signatures in their Arabic outputs, with domain-specific characteristics that vary significantly between different contexts. Based on these insights, we developed BERT-based detection models that achieved exceptional performance in formal contexts (up to 99.9% F1-score) with strong precision across model architectures. Our cross-domain analysis confirms generalization challenges previously reported in the literature. To the best of our knowledge, this work represents the most comprehensive investigation of Arabic machine-generated text to date, uniquely combining multiple prompt generation methods, diverse model architectures,

---

[*]Corresponding author

  *Email address:* `moataz.ahmed@kfupm.edu.sa` (Maged S. Al-Shaibani, Moataz Ahmed)

and in-depth stylometric analysis across varied textual domains, establishing a foundation for developing robust, linguistically-informed detection systems essential for preserving information integrity in Arabic-language contexts.

---

## 1. Introduction

The landscape of natural language processing has undergone a revolutionary transformation with the emergence of increasingly sophisticated Large Language Models (LLMs). Models such as the GPT family [1], Claude [2], and open-source alternatives like Llama [3], GPT-Neo [4], Phi [5], and Command-R [6] have achieved unprecedented capabilities in text generation. These models now produce content that is increasingly indistinguishable from human-authored text [7, 8], demonstrating remarkable proficiency across domains, styles, and complexity levels. This leap in capability stems from their enormous scale, sophisticated architectures, and exposure to vast training datasets, enabling them to capture sophisticated patterns in human language and generate contextually appropriate, coherent responses.

While these technological advancements create valuable opportunities for automation and content creation, they simultaneously present profound challenges for information integrity and verification [9]. The ability of modern LLMs to generate highly coherent and contextually relevant text has raised serious concerns across multiple sectors. Educational institutions now face unprecedented challenges in maintaining academic integrity as students gain access to sophisticated tools capable of generating essays, research papers, and even technical content [10, 11]. News organizations and publishing platforms struggle with content verification as the barriers between authentic and synthetic content blur. Perhaps most concerning, these models enable the creation of sophisticated disinformation at scale, potentially undermining public discourse and facilitating targeted influence campaigns. Unlike earlier generations of language models that produced text with relatively obvious patterns, modern LLMs generate sophisticated content that closely mimics human writing styles, including advanced arguments, creative expressions, and domain-specific terminologies [12].

The challenge of machine-generated text detection becomes particularly acute in multilingual contexts, especially for languages with rich linguistic traditions but limited computational resources. While significant research has focused primarily on English, languages such as Arabic present unique challenges that remain under-explored. The Arabic language space has witnessed notable developments with specialized models such as Jais [13], AceGPT [14], and ALLaM (Arabic Large Language Model) [15], yet the field of Arabic machine-generated text detection lags behind its English counterpart [16, 17]. Recent studies have highlighted limitations in existing Arabic machine-generated detection systems when processing Arabic scripts, particularly in handling diacritized content and maintaining consistent performance across different Arabic text variants [16].

Current research has identified several critical challenges in detecting machine-generated text, including the need for robust cross-model generalization, the importance of considering various generation strategies, and the difficulty of maintaining detection accuracy across different languages and domains [18]. The emergence of methods to bypass detection systems, such as paraphrasing and hybrid human-AI content [19], further complicates this task. Despite these challenges, no comprehensive study has explored Arabic machine-generated text detection across multiple domains, generation strategies, and model architectures, a gap our work addresses.

To the best of our knowledge, this work constitutes the first systematic and comprehensive study of Arabic machine-generated text detection, covering multiple domains including academic prose, and social media discourse. We employ a systematic approach that combines multiple data sources, generation strategies, LLM architectures, and analytical methods. We also made our work publicly available[1]. Specifically, we make the following contributions:

1. **Multi-dimensional stylometric comparative analysis:** We conduct the first comprehensive stylometric analysis of human-written versus machine-generated Arabic text across different domains, examining word frequency distributions, semantic metrics, and statistical patterns. This analysis reveals distinctive linguistic signatures that characterize machine-generated Arabic text despite its human-like qualities.
2. **Multi-prompts generation framework:** We systematically evaluate multiple generation methods, including title-only generation, content-aware generation, and text refinement approaches, across four distinct LLM architectures (ALLaM, Jais, Llama 3.1, and GPT-4). This framework provides insights into how different generations' approaches affect linguistic patterns and detectability. The total number of generated samples out of our framework is 11.7k.
3. **Machine-generated detection systems:** Based on our linguistic analysis, we develop and evaluate BERT-based detection models that achieve notable performance (up to 99.9% F1-score) in formal contexts with strong cross-model generalization capabilities. These models demonstrate the feasibility of reliable Arabic text detection despite the sophisticated generation capabilities of modern LLMs.

---

[1] https://github.com/KFUPM-JRCAI/arabic-text-detection

The rest of this paper is organized as follows: Section 2 provides a comprehensive review of related work in machine-generated text analysis and detection. Section 3 details the datasets used in our investigation, focusing on Arabic academic abstracts and social media content. Section 4 presents our methodology, including text generation methods and detection approaches. Section 5 presents a detailed linguistic analysis of machine-generated Arabic text compared to human-written content. Section 6 covers detection results and analysis. Finally, Section 7 concludes the work with a summary of our contributions and their implications for the field of machine-generated text detection in Arabic contexts.

## 2. Literature Review

Research on LLM-generated text detection spans multiple approaches: training-based methods using supervised learning, zero-shot methods requiring minimal training, and watermarking techniques embedding detectable patterns during generation. These address concerns about LLM misuse while acknowledging the increasing difficulty in distinguishing between human and machine content [9], [20].

DetectGPT [7] leverages the observation that machine-generated text occupies negative curvature regions of model log probabilities, achieving 0.95 AUROC for fake news detection without requiring separate classifiers or datasets. Fast-DetectGPT [8] improved on this by introducing conditional probability curvature, reducing computational cost by 340 times while improving accuracy by 75% in both white-box and black-box settings.

For academic integrity, CHEAT [10] provides 35,000+ synthetic academic abstracts for developing ChatGPT detection methods, revealing increased difficulty when human involvement exists. Similarly, Liang et al. [11] analyzed nearly one million scientific papers, estimating LLM-modified content at population level rather than detecting individual instances, finding Computer Science papers with the highest LLM usage rate (17.5%).

Stylometric analysis, explored in authorship atribution research [21, 22], can be used as an effective tool to distinguish machine-generated text [23]. Herbold et al. [24] showing ChatGPT essays rated higher in quality yet exhibiting different linguistic characteristics in lexical diversity and sentence complexity. For short-form content, Kumarage et al. [25] developed "stylometric change point agreement" to identify AI-generated tweets by analyzing stylistic timeline changes, building on earlier work proposed by Feng et al. [26] and Takahashi and Tanaka-Ishii [27].

The HC3 dataset [28] pioneered ChatGPT detection with 40K human/ChatGPT answers across various domains, revealing distinctive AI stylometric patterns. HC3 Plus [29] later demonstrated that detection methods struggle with semantic-preserving transformations like summarization.

Muñoz-Ortiz et al. [12] quantitatively compared human-written news text with six LLMs, finding humans exhibit more varied sentence lengths, greater vocabulary diversity, distinct dependency types, shorter constituents, and more optimized dependency distances. Humans express stronger negative emotions, while LLMs show more objective language with increasing toxicity as model size grows.

To address paraphrasing-based detection evasion [28, 29], Koike et al. [19] introduced a framework improving detector robustness through adversarial learning, achieving up to 41% F1-score improvement when detecting adversarially generated texts.

For real-world applications, MAGE [30] evaluated text from various domains and multiple LLMs, showing detection methods work well in specific domains but deteriorate significantly with diverse texts or out-of-distribution scenarios, as linguistic differences between human and machine text converge.

For Arabic specifically, Alshammari and El-Sayed [16] proposed a benchmark for evaluating AI detectors on Arabic text, highlighting limitations in handling diacritics. Current detectors like GPTZero [31] struggled with Arabic text detection. Alshammari et al. [17] developed an Arabic AI detector using AraELECTRA [32] and XLM-R [33], achieving 98.4% accuracy compared to GPTZero's 62.7%.

These findings underscore the complexity of LLM-generated text detection and highlight research directions for low-resource languages like Arabic, including developing robust cross-domain and cross-lingual detection methods, creating evaluation frameworks, and addressing hybrid human-AI content detection.

## 3. Datasets

We studied this problem by focusing on two primary domains: Arabic academic abstracts for scholarly writing and social media reviews for informal content. This section details the collection and construction of the datasets we utilized from these domains.

### 3.1. Arabic Academic Abstracts Dataset

To study academic writing, we built a dataset of Arabic academic papers and their abstracts from the Algerian Scientific Journals database, a platform containing a multitude of Arabic academic papers across diverse domains [2]. We utilized web scraping to collect metadata and PDFs from over 60,000 papers, including titles, journal names, volumes, and publication dates. We then filtered papers to those published between 2010 and 2022 to avoid potential AI-generated content.

The data processing presented several unique challenges. First, these papers have abstracts written in at least on of the following languages: Arabic, English, or French. Unfortunately, the site dumps these abstracts in a single text block. Hence, we wrote custom segmentation scripts that use statistical analysis and rule-based methods to segment them. We further employed language detection tools to distinguish between language segments and implemented validation rules to verify consistency at abstract boundaries.

Second, for generation methods that require paper content extraction, we faced additional challenges related to Arabic text extraction from PDF. Unlike English papers that usually provide LaTeX source files, we worked directly with PDFs, employing PyPDF2[3] for direct text extraction while avoiding OCR due to known complications with Arabic script. The extracted text posed significant formatting challenges due to Arabic's cursive script, requiring extensive preprocessing including Unicode normalization, language verification, removal of duplicated headers and footers, and careful standardization of whitespace without disrupting meaning[4].

---

[2] https://asjp.cerist.dz/
[3] https://pypi.org/project/PyPDF2/

After applying our filtering criteria and processing pipeline, we curated a dataset of 3,000 papers with their abstracts. 1,619 of these papers had Arabic-only abstracts while 1,381 had Arabic and English abstract pairs. In this dataset, abstract lengths in terms of the number of words range from 75 to 294 words, with an average of 120 words per abstract.

*3.2. Social Media Reviews Dataset*

To explore detection in casual writing contexts, we built a dataset from two prominent Arabic review collections: BRAD (Book Reviews in Arabic Dataset) [34] collected from Goodreads.com and HARD (Hotel Arabic Reviews Dataset) [35] collected from Booking.com. Both datasets primarily contain Modern Standard Arabic text, with loose language and informal tone.

We tried to obtain longer-form reviews suitable for meaningful linguistic analysis. From BRAD, we selected 3,000 reviews containing between 724-1,500 words per review, leveraging the naturally longer format of book reviews. From HARD, due to the typically shorter nature of hotel reviews, we extracted 500 reviews ranging from 150-614 words. To ensure data quality, we applied several preprocessing steps, including removal of special characters and non-printable text, normalization of Arabic text through tatweel removal, and standardization of repeated punctuation marks (limiting repetitions to a maximum of 3). The final dataset consists of 3,500 reviews with an average length of 890.6 words, as shown in Table 1.

## 4. Methodology

This section outlines our approach to studying Arabic LLMs generated text. We investigated this problem from two main aspects: first, we generated text from AI with various generation methods across multiple models. Second, we developed BERT-based detectors to identify machine-generated text, studying various scenarios like multi-class detection, and cross-model detection. Furthermore, we performed further analysis of human text vs

---

[4]The extraction notebook can be found here: https://github.com/KFUPM-JRCAI/arabs-dataset/blob/main/notebooks/explore_and_extract.ipynb

| Metric | Value |
| --- | ---: |
| Total Samples | 3,500 |
| Maximum Length (words) | 1,500 |
| Minimum Length (words) | 150 |
| Average Length (words) | 890.6 |
| BRAD Samples | 3,000 |
| HARD Samples | 500 |
| BRAD Length Range | 724-1,500 |
| HARD Length Range | 150-614 |

**Table 1:** Social Media-extracted Dataset Statistics.

AI text in a dedicated section of stylometric analysis (Section 5). Figure 1 overviews our framework.

*4.1. Text Generation Strategies*

To analyze the characteristics of machine-generated Arabic text, we generated text across different models and domains with different strategies. This allowed us to study how these different factors influence linguistic patterns and detectability.

*4.1.1. Academic Abstracts Generation*

For academic abstracts, we generated abstracts using three different methods described as follows.

First, we generate abstracts from only the paper title. The prompt asked the LLM to generate abstracts with an approximate length of 100-150 words, trying to match the human-written abstracts' size. In this method, we are experimenting with generation in a free-form approach with minimal input.

Second, we generate abstracts from both the title and the paper content. We happened to the prompt an appropriate portion of the paper content. This limit was 400 words to accommodate models' contexts' length (Jais and ALLaM max context window is 4096), given the special Arabic characters (diacritics, for instance) and the degraded quality of PDF extraction, resulting in a more fine-grained tokenization. The prompt addressed unique challenges of PDF-extracted Arabic text, including guidelines for handling extraction artifacts, character segmentation issues, and diacritical mark distortions. Since it was difficult to remove the abstract from the extracted

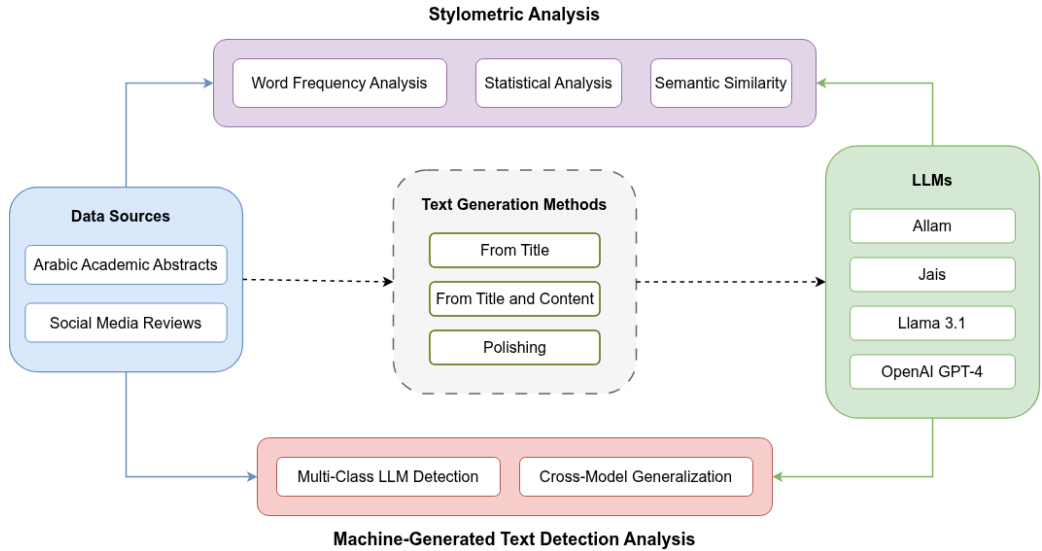

**Figure 1:** General overview of our research pipeline. The study integrates multiple data sources with diverse text generation strategies across four different LLMs. The generated texts undergo stylometric analysis. Detection models are then developed and evaluated through cross-model generalization and multi-class identification.

PDF, models were explicitly instructed in the prompt to disregard any existing abstracts in the extracted content. This approach simulates an abstract summarization approach of the paper content.

Finally, we generate abstracts by asking the model to polish the currently existing human abstract, improving linguistic and stylistic elements while maintaining the original content. We are interested in observing how models will respond in this proofread kind of generation compared to the other aforementioned methods.

Additionally, we filtered the generated abstracts, removing invalid ones containing error messages or those falling below a threshold of 30-word count. After this filtration step, the final common set across all models decreased from the initial collection (3k samples), with the most significant reduction observed in the Title-Content generation approach due to the fact that Jais faced challenges in generating abstracts with this type of generation (reduced to 2,575 abstracts from original counts), representing approximately a 15% filtration rate. Other generation types are marginally affected, with approximately 120 samples dropped. The dataset is publicly available[5]. The prompts used to generate these samples can also be found in the associated repository of this work[6].

*4.1.2. Social Media Post Generation*

For social media posts, we experimented with the polishing generation approach, emphasizing preserving any dialectal expressions and diacritical marks in the original post, recognizing the importance of these elements in social media content. Models were instructed to use vocabulary closely aligned with the original text while correcting grammatical or spelling errors without altering the fundamental writing style. The prompt included guidelines for maintaining text coherence while preserving the informal nature of social media writing. We emphasized maintaining approximate word counts close to the original human posts to ensure generated content remained comparable in scope and depth. After generation, we applied some filtering by removing invalid generated posts, setting a minimum threshold of 50 words per post, and dropping duplicated samples, if any. The resulting final set after this filtration becomes 3318 samples. The dataset is publicly available[7]. The prompts used to generate these samples can also be found in the associated repository of this work[8].

*4.1.3. LLM Selection*

We select models with variety to evaluate the generation capabilities across a diverse range of architectures and specialties. We deliberately included both Arabic-specialized models and general-purpose multilingual systems to assess how language specialization affects generation quality and detectability. For the model size, we tried to cover both extremes, small and large sizes, as well as open and closed source models. Table 2 presents these models.

---

[5] https://huggingface.co/datasets/KFUPM-JRCAI/arabic-generated-abstracts

[6] Inspect the prompts in this notebook (all models got the same prompts) https://github.com/KFUPM-JRCAI/arabic-text-detection/blob/main/notebooks/Arabic_synthetic_dataset_generation/AbstractsDataset/allam.ipynb

[7] https://huggingface.co/datasets/KFUPM-JRCAI/arabic-generated-social-media-posts

[8] Inspect the prompts in this notebook (all models got the same prompts)https://github.com/KFUPM-JRCAI/arabic-text-detection/blob/main/notebooks/Arabic_synthetic_dataset_generation/SocialMediaDataset/allam.ipynb

| Model | Size (B) | domain | Source |
|---|---|---|---|
| ALLaM [15] | 7 | Arabic-focused | Open |
| Jais [13] | 70 | Arabic-focused | Open |
| Llama 3.1 [36] | 70 | General | Open |
| OpenAI GPT-4 [37] | – | General | Closed |

**Table 2:** Large Language Models Used to generate text.

*4.2. Detection Methodology*

Based on the linguistic analysis of generated content, we developed a detection approach using fine-tuned transformer models. We employed XLM-RoBERTa, a multilingual BERT-based model featuring 279M parameters, 12 attention heads, and a 512-token context window. This model was pre-trained on 2.5TB of filtered CommonCrawl data spanning 100 languages. We used Huggingface Transformers and PyTorch-Lightning to fine-tune the model with early stopping of 3 consecutive evaluation if not improvement observed, conducting 4 evaluations per epoch with a batch size of 64.

For detection analysis, we first investigate cross-model generalization by training detection models on content from one LLM and testing their performance on content generated by different LLMs, assessing the models' ability to generalize across various LLMs. Second, we explore multi-class LLM identification, where models are trained not only to detect machine-generated content but also to identify the specific LLM responsible for generating it.

For each experiment, we evaluated performance using standard classification metrics including accuracy, precision, recall, and F1-score to provide comprehensive assessment of model effectiveness.

## 5. Stylometric Analysis of Machine-Generated Arabic Text

This section presents the stylometric analysis we applied, comparing human-written vs AI text. We studied both of these texts from basic statistics, their Zipfian distribution, and syntactic and semantic similarity. We applied this analysis to our two datasets: academic abstracts and social media.

*5.1. Academic Abstracts Analysis*

This section presents our stylometric analysis applied to the Academic abstracts dataset. This analysis reveals notable differences between human and machine-generated text across all aspects we studied and even among the generation methods used.

*5.1.1. Length and Statistical Analysis*

We start by examining the basic statistical properties of generated academic abstracts compared to those of human-written ones. As shown in Table 3, the human average length is 120 words per abstract, while AI-generated texts showed varying patterns across different generation methods. OpenAI was the closest to the human word length, particularly in title-only generation (123.3 words). In contrast, Jais produced notably shorter abstracts in title-only (62.3 words) and polishing (68.5 words) generations, while showing better alignment in content-based generation (105.7 words). Llama and ALLaM abstracts generally fell between these extremes.

| Generation | ALLaM | Jais | Llama | OpenAI | Human |
|---|---|---|---|---|---|
| Title-Only | 77.2 | 62.3 | 99.9 | 123.3 | |
| Title-Content | 95.3 | 105.7 | 103.2 | 113.9 | 120 |
| Abstract Polish | 104.3 | 68.5 | 102.3 | 165.1 | |

**Table 3:** Statistical Analysis of generated abstract word counts.

From these results, we can see that the generation method notably influenced the abstract length for all models. Title-only generation, which represents the free-form generation, typically produced shorter abstracts (except for OpenAI), while content-based generation resulted in more consistent lengths across models. Abstract polishing showed the most variation, with OpenAI significantly exceeding human length averages (165.1 words) while Jais produced the shortest abstracts (68.5 words).

*5.1.2. Top Frequent Words*

To explore the linguistic characteristics and diversity of the generated text, we analyzed the most frequent words used in both human and LLM-generated abstracts. Table 4 presents the top 10 frequent words across human and AI-generated texts for different generation types with their frequencies. Before generating this table, we preprocessed texts by removing punctuations and lower casing to cover any latin script words available in the text. We also removed stop words as we believe that such words may provide misleading insights.

**Table 4:** Most Frequent Words of Human Abstracts VS LLMs.

**Legend:** #: Rank. <span style="background-color:#c8e6c9">Human Match</span> Words in both human and LLM texts. <span style="background-color:#ffcdd2">Shared LLM</span> Words common across all LLMs. <span style="background-color:#ffcc80">Single LLM</span> Words unique to one LLM. <span style="background-color:#c5cae9">Cross LLM</span> Words in multiple but not all LLMs. <span style="background-color:#dce775">Human Unique</span> Words unique to human texts. **Stable Position** Words with similar rank in 3 columns.

| # | Human | ALLaM | OpenAI | Jais | Llama |
|---|---|---|---|---|---|
| *Title-only Generation* | | | | | |
| 1 | الدراسة (1972) | الدراسة (6533) | الدراسة (6039) | الورقة (2703) | البحث (7594) |
| 2 | خلال (1563) | تهدف (2598) | النتائج (3075) | النتائج (2640) | الدراسة (3095) |
| 3 | البحث (819) | المتوقع (2143) | تحليل (2973) | البحثية (2318) | خلال (2899) |
| 4 | الجزائر (718) | تحليل (2046) | الورقة (2306) | تدرس (2004) | يهدف (2129) |
| 5 | العربية (639) | خلال (1948) | تهدف (2063) | الدراسة (1889) | دراسة (2079) |
| 6 | الجزائري (582) | الضوء (1838) | خلال (1997) | خلال (1371) | منهجية (1770) |
| 7 | أهم (551) | النتائج (1791) | البحث (1676) | التركيز (1330) | النتائج (1661) |
| 8 | تم (516) | البحث (1764) | منهجية (1584) | تشير (1326) | فهم (1458) |
| 9 | الباحث (482) | استخدام (1711) | تعزيز (1511) | يتم (1242) | تعزيز (1401) |
| 10 | العلمي (475) | لتحقيق (1447) | تعتمد (1456) | تحليل (1232) | تحليل (1368) |
| *Title-and-Content Generation* | | | | | |
| 1 | الدراسة (1972) | البحث (5397) | البحث (4313) | البحث (2896) | البحث (6918) |
| 2 | خلال (1563) | الدراسة (2739) | الدراسة (3666) | الدراسة (2022) | أهمية (2059) |
| 3 | البحث (819) | يتناول (1763) | خلال (1605) | يمكن (1376) | الدراسة (1507) |
| 4 | الجزائر (718) | خلال (1058) | أهمية (1373) | خلال (1221) | خلال (1384) |
| 5 | العربية (639) | يهدف (1007) | الضوء (994) | يتم (1204) | دراسة (1339) |
| 6 | الجزائري (582) | الجزائر (862) | تحليل (882) | تم (1126) | يظهر (1275) |
| 7 | أهم (551) | دراسة (800) | يتناول (859) | الجزائر (946) | يهدف (1139) |
| 8 | تم (516) | يشير (761) | تعزيز (855) | النص (935) | الجزائر (991) |
| 9 | الباحث (482) | تهدف (732) | تهدف (845) | بشكل (847) | الضوء (942) |
| 10 | العلمي (475) | تحليل (729) | الجزائر (820) | التركيز (796) | يعد (901) |
| *Polishing Abstracts Generation* | | | | | |

**Table 4:** Most Frequent Words of Human Abstracts VS LLMs.

**Legend:** #: Rank. <mark style="background-color:#cdeac0">Human Match</mark> Words in both human and LLM texts. <mark style="background-color:#f9cfd4">Shared LLM</mark> Words common across all LLMs. <mark style="background-color:#f7c08a">Single LLM</mark> Words unique to one LLM. <mark style="background-color:#c9c9f0">Cross LLM</mark> Words in multiple but not all LLMs. <mark style="background-color:#d8e28a">Human Unique</mark> Words unique to human texts. **Stable Position** Words with similar rank in 3 columns.

| # | Human | ALLaM | OpenAI | Jais | Llama |
|---|---|---|---|---|---|
| 1 | الدراسة (1972) | الدراسة (3270) | الدراسة (3852) | الدراسة (1723) | البحث (3059) |
| 2 | خلال (1563) | البحث (2555) | البحث (2879) | خلال (1308) | خلال (2252) |
| 3 | البحث (819) | خلال (2283) | خلال (2524) | تم (955) | الدراسة (2043) |
| 4 | الجزائر (718) | دراسة (1169) | تعزيز (1427) | التركيز (918) | أهمية (1170) |
| 5 | العربية (639) | تحليل (1147) | تحليل (1280) | يتم (892) | التركيز (920) |
| 6 | الجزائري (582) | تهدف (1119) | الضوء (1166) | تدرس (789) | دراسة (846) |
| 7 | أهم (551) | الجزائر (902) | أهمية (1147) | الجزائر (701) | الجزائر (812) |
| 8 | تم (516) | الضوء (890) | بشكل (1124) | البحث (669) | حول (781) |
| 9 | الباحث (482) | تم (828) | الجزائر (1091) | بشكل (514) | يبرز (764) |
| 10 | العلمي (475) | بشكل (795) | تهدف (1026) | العربية (489) | الضوء (729) |

From this table, we can notice that human-authored abstracts exhibited broader vocabulary diversity, with 40% of their top frequent words being unique domain-specific academic terms (e.g., الباحث، العلمي) or region-specific (الجزائري). Additionally, LLMs showed higher repetition rates with most frequent words exceeding 3,000 occurrences versus under 2,000 in human texts. Each LLM (except OpenAI) demonstrated unique linguistic signatures with at least four distinctive words absent from both human texts and other LLM outputs, though significant commonality existed in their most frequent words across generation types. The generation approach seems to influence the repetition patterns, with title-only generation showing the highest repetition frequency and polishing-based generation most closely resembling human writing patterns.

To gain further insights from this frequency analysis, we plot the top 100 frequent words frequencies of both human-authored and AI-generated abstracts. Figure 2 depicts this analysis across different LLMs and generation methods. From this figure, we can notice that human-authored ab-stracts (black solid line) demonstrated a relatively smooth power-law decay, particularly evident in the range of ranks 1-50. Additionally, all models consistently overused high-frequency words (ranks 1-5) compared to human writing, with their markers appearing notably above the human reference line in this region. In the mid-frequency range (ranks 5-20), models showed varying alignment with human patterns, though most exhibited a tendency to underutilize vocabulary in this range. The most revealing differences happened in the long tail (ranks 50-100), where all AI models demonstrated a substantially steeper frequency drop-off compared to human texts.

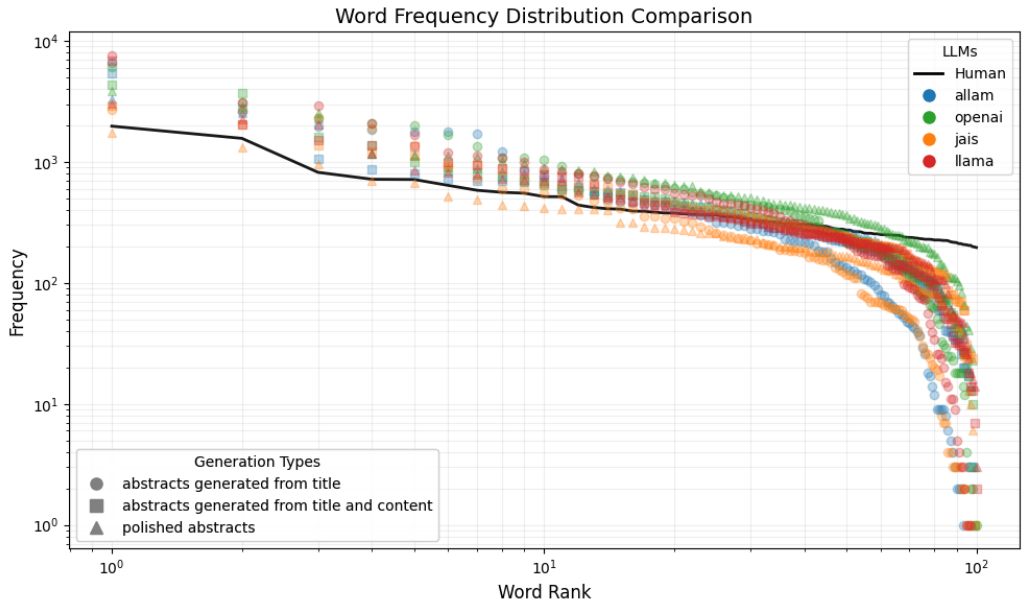

**Figure 2:** Arabic abstracts top 100 human vs LLMs words frequency.

*5.1.3. Syntactic and Semantic Similarity*

To assess textual similarity between generated and human abstracts, we employed multiple reference-based metrics examining different levels of linguistic correspondence that cover BLEU [38], METEOR [39], ROUGE-L [40] for syntactic-based similarity, and BERTScore [41] for semantic-based similarity. Table 5 presents these results across all models and generation methods.

From this table, a performance improvement can be noticed from title-only generation to polishing approaches, with each additional context level enhancing generation quality. The polishing approach yielded the highest similarity scores. Each model exhibited distinctive strengths: LLaMA demonstrated superior performance across most metrics, particularly excelling in polishing approaches with the highest BLEU (20.42), METEOR (37.58), and BERTScore (80.64) values. Interestingly, Jais diverged from this pattern, performing best in the Title+Content approach with BLEU (12.21). The consistently elevated BERTScore values (67.90-80.64) compared to syntactic-level metrics suggest that generated abstracts preserved essential semantic meaning even when exact wording differed substantially. It is also interesting to note the low syntactic level values happening across multiple models and generation methods, especially in the free-form generation (title-only generation). This indicates that these models tend not to extensively utilize Arabic abstracts technical language.

| Model | Method | BLEU | METEOR | ROUGE-L | BERTScore |
|---|---|---|---|---|---|
| ALLaM | Title-only | 1.80 | 9.37 | 3.33 | 67.90 |
| | Title+Content | 11.75 | 25.32 | 14.00 | 75.53 |
| | Polishing | 15.64 | 32.94 | 23.65 | 78.79 |
| OpenAI | Title-only | 1.61 | 10.10 | 3.08 | 68.18 |
| | Title+Content | 6.06 | 19.44 | 15.75 | 73.99 |
| | Polishing | 6.40 | 24.65 | 24.17 | 75.59 |
| Jais | Title-only | 0.97 | 8.28 | 2.86 | 69.32 |
| | Title+Content | 12.21 | 26.00 | 14.66 | 75.73 |
| | Polishing | 8.81 | 24.42 | 21.62 | 77.65 |
| LLaMA | Title-only | 2.31 | 10.20 | 2.86 | 68.60 |
| | Title+Content | 14.02 | 27.19 | 12.16 | 75.71 |
| | Polishing | 20.42 | 37.58 | 22.38 | 80.64 |

**Table 5:** Automatic Evaluation Results Across Different Generation Methods (All Metrics on 0-100 Scale).

## 5.2. Social Media Content Analysis

We extend our analysis framework applied to the abstracts dataset to the social media dataset. Similarly to the abstracts dataset, we observed many interesting yet distinguishing patterns comparing human vs. AI text.

### 5.2.1. Length and Statistical Analysis

Table 6 shows that human-authored posts maintained substantially longer lengths, averaging 867.4 words, while LLM outputs were consistently shorter.

This length reduction was most notable with Llama (225.3 words on average), followed by Jais (305.3 words), representing approximately 25% and 35% of human length, respectively. Due to this notable variation, it becomes interesting to see how these LLMs will behave at their extreme generation. We can notice that while human posts reached a maximum of 1,546 words, ALLaM and OpenAI occasionally produced even longer content (2,705 and 1,761 words, respectively). In contrast, Jais and Llama maintained much stricter length constraints, with maximum values of only 409 and 443 words, respectively.

| Word Count | ALLaM | OpenAI | Jais | LLaMA | Human |
|---|---|---|---|---|---|
| Min | 50.0 | 76.0 | 53.0 | 73.0 | 135.0 |
| Max | 2705.0 | 1761.0 | 409.0 | 443.0 | 1546.0 |
| Avg | 627.4 | 449.5 | 305.3 | 225.3 | 867.4 |

**Table 6:** Statistical Analysis of Generated Social Media Post Word Counts.

*5.2.2. Top Frequent Words Analysis*

Table 7 presents the top 10 frequent words for human and AI-generated texts. Before generating this table, we applied similar preprocessing steps applied in the abstracts analysis. From this table, we can notice that human-written posts demonstrate greater diversity, with more balanced and natural word distributions. Frequencies gradually decline from about 7,977 occurrences for the top-ranked word to 2,470 for the tenth-ranked word. In contrast, except for ALLaM, LLM outputs show steeper frequency drops, particularly evident in Llama's output where the top word appears 2,905 times while the tenth-ranked word occurs only 714 times. Note that, referring to table 6, human posts are longer than LLMs posts. ALLaM's text in this context behaves most similarly to human text in terms of word frequencies, aligning with our previous observation that it produced the closest approximation to human post lengths. On the other hand, each LLM exhibits unique linguistic signatures, with three unique words for both Llama and Jais and one word each for OpenAI and ALLaM. Interestingly, Llama introduces domain-specific terms like الفندق (hotel), capturing vocabulary from the hotels' reviews subset that other LLMs missed.

Figure 3 presents the top 100 frequent words with their frequencies of both AI and human text. From this figure, the human-authored posts (black solid

**Legend:** <span style="background-color:#c8e6c9">Human Words</span> Words in human texts column. <span style="background-color:#ffcdd2">Shared LLM</span> Words common across all LLMs. <span style="background-color:#ffe0b2">Single LLM</span> Words unique to one LLM. <span style="background-color:#d1c4e9">Cross LLM</span> Words in multiple but not all LLMs. <span style="background-color:#f0f4c3">Human Unique</span> Words unique to human texts. **Stable Position** Words with similar rank in 3 columns.

| # | Human | ALLaM | OpenAI | Jais | Llama |
|---|---|---|---|---|---|
| 1 | الرواية (7977) | الرواية (7378) | الرواية (5699) | الرواية (4828) | الكتاب (2905) |
| 2 | الله (7370) | الكتاب (6131) | الكتاب (4875) | الكتاب (4386) | الرواية (2798) |
| 3 | الكتاب (6659) | الله (5381) | الله (3868) | الكاتب (2282) | الله (1652) |
| 4 | الكاتب (4351) | الكاتب (4025) | الكاتب (2711) | يمكن (2252) | الفندق (1336) |
| 5 | الناس (3452) | الحياة (2743) | الحياة (2190) | الله (2124) | الكاتب (1297) |
| 6 | الحياة (3390) | الناس (2477) | الإنسان (2069) | بشكل (1730) | رواية (1101) |
| 7 | نفسه (2944) | الإنسان (2417) | الحب (1817) | الفندق (1516) | كتاب (859) |
| 8 | العالم (2585) | يمكن (2203) | يمكن (1752) | يجب (1481) | الحياة (814) |
| 9 | يمكن (2522) | بشكل (2126) | بشكل (1693) | الحياة (1480) | الإنسان (744) |
| 10 | الحب (2470) | نفسه (2037) | الناس (1689) | خلال (1345) | الناس (714) |

**Table 7:** Most Frequent Words in Social Media Posts: Human VS LLMs.

line) exhibit a classic Zipf-like distribution with a relatively smooth power-law decay across all ranks. LLM-generated texts, while exhibiting similar trends to the human text, show a notable vertical shift at the long-tail due to the posts' lengths variation, having lower overall frequencies compared to human text. Note that we generated posts by polishing the human posts. For model-specific analysis, ALLaM shows the closest alignment to human writing (particularly in mid-frequency ranks 4-20), while OpenAI and Jais exhibit similar patterns with steadier frequency declines. Llama demonstrates the most distinctive distribution with steeper drop-offs and generally lower frequencies across most ranks.

*5.2.3. Semantic Similarity Analysis*

Similar to the abstracts dataset, we assess the semantic preservation of the social media dataset using the same similarity metrics. Table 8 presents these results across all models.

Semantic analysis reveals model-specific features in preserving different aspects of original content. ALLaM demonstrates superior performance in syntactic-level metrics (BLEU: 45.10, METEOR: 54.91, ROUGE-L: 42.26), preserving exact phrasing and structural elements, consistent with its human-like length and word distribution patterns, and the second in BERTScore. Conversely, LLaMA achieves the highest BERTScore (95.05) despite producing the shortest text and lower surface-level scores, indicating remarkable semantic preservation at a contextual level while using different wording and writing structures. These high BERTScores across all models (81.71-95.05) suggest that while LLMs modify length and exact phrasing significantly, they maintain the essential meaning of reference posts. It is also interesting to note the generally higher results in this type of text compared to the academic abstracts dataset.

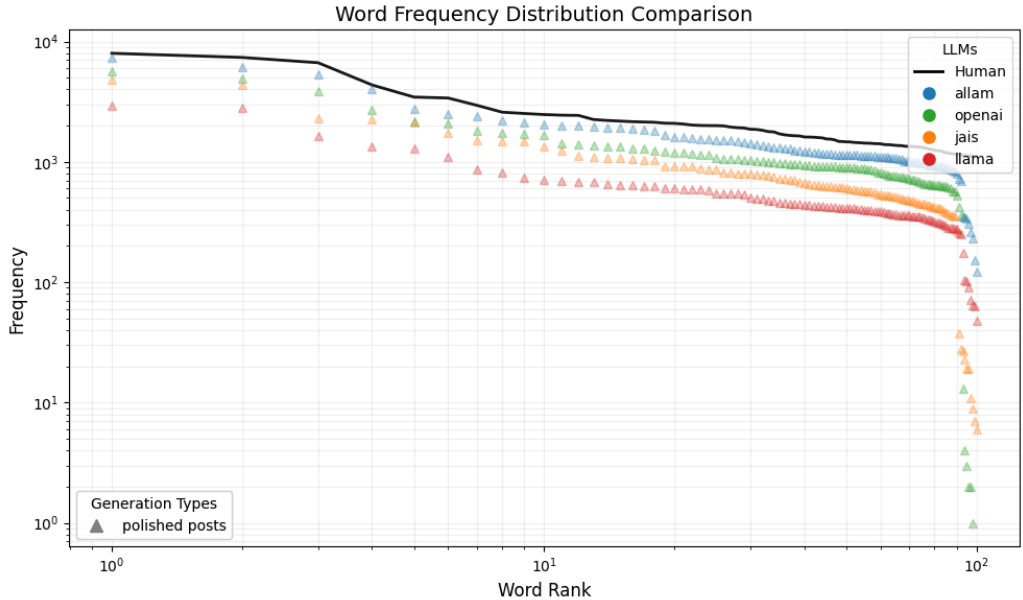

**Figure 3:** Arabic social media top 100 human vs LLMs words frequency.

## 6. Detection Results and Analysis

This section presents our findings on detecting machine-generated Arabic content across the domains, models, and generation methods we employed in this study. Our results demonstrate both the effectiveness of our detection approach and its varying performance characteristics across formal academic and informal social media contexts.

| Model | BLEU | METEOR | ROUGE-L | BERTScore |
|---|---|---|---|---|
| ALLaM | 45.10 | 54.91 | 42.26 | 86.36 |
| OpenAI | 11.30 | 27.18 | 37.14 | 82.31 |
| Jais | 5.58 | 20.38 | 22.22 | 81.71 |
| LLaMA | 4.24 | 26.56 | 26.15 | 95.05 |

**Table 8:** Semantic Similarity Scores for Generated Social Media Posts (All Metrics on 0-100 Scale).

### 6.1. Arabic Academic Abstracts Detection

Our detection experiments with Arabic academic abstracts reveal strong performance across various detection scenarios. The formal nature of academic writing appears to create distinctive patterns that facilitate a reliable detection of machine-generated content, even across different model architectures. Note that the dataset with which we experimented is imbalanced, as we included human-generated text along with all samples from each generation method we applied. So, the dataset has roughly a 1:3 ratio of human vs machine-generated text. Throughout the experiments in this subsection, we split the dataset into a 75%, 15%, and 15% train, validation, and test splits unless specified otherwise.

#### 6.1.1. Cross-Model Generalization

In this experiment, we want to study the effect of LLM architecture on the detection process. We split each LLM's text into train and test splits. We then train our BERT-based model on a dataset compiled from an LLM train set and human text, then test it on all LLM test sets.

Table 9 presents the detailed results across multiple classification metrics. This analysis reveals several noteworthy patterns. First, both ALLaM-trained and OpenAI-trained detectors achieve 100% precision across all test sets (highlighted in cyan), indicating that when they identify content as AI-generated, they are never wrong. This perfect precision reflects the presence of clear, distinctive features in LLM-generated Arabic academic text from these models. Others achieved competitive performance (more than 99%). We can also note that all models excel at identifying their own outputs (with F1-scores above 99.5%), demonstrating the distinctiveness of each model's signature. However, the varying recall values, particularly OpenAI's struggle with Jais-generated content (76.57%), indicate that detectors sometimes miss AI content from other models. This suggests each AI has unique stylistic patterns that other detectors might not recognize. While all detectors generally achieved competitive performance on F1 score, the most robust cross-model detection comes from Jais and Llama-trained detectors, which generalize well across architectures.

| Metric | Train → Test | ALLaM | Jais | Llama | OpenAI |
|---|---|---|---|---|---|
| Accuracy | ALLaM | **99.94** | 94.55 | 90.51 | 96.19 |
|  | Jais | 99.88 | **99.53** | 97.66 | 99.88 |
|  | Llama | 99.82 | 99.00 | **99.36** | 99.77 |
|  | OpenAI | 91.56 | 82.66 | 93.44 | **99.94** |
| Precision | ALLaM | 100.0 | 100.0 | 100.0 | 100.0 |
|  | Jais | 99.83 | 99.83 | 99.83 | 99.83 |
|  | Llama | 99.83 | 99.83 | 99.83 | 99.83 |
|  | OpenAI | 100.0 | 100.0 | 100.0 | 100.0 |
| Recall | ALLaM | **99.92** | 92.66 | 87.26 | 94.80 |
|  | Jais | 100.0 | **99.51** | 97.02 | 100.0 |
|  | Llama | 99.91 | 98.81 | **99.29** | 99.84 |
|  | OpenAI | 88.59 | 76.57 | 91.14 | **99.92** |
| F1-Score | ALLaM | **99.96** | 96.11 | 93.07 | 97.28 |
|  | Jais | 99.91 | **99.66** | 98.38 | 99.91 |
|  | Llama | 99.87 | 99.30 | **99.55** | 99.83 |
|  | OpenAI | 93.80 | 86.39 | 95.28 | **99.96** |

**Table 9:** Cross-Model Detection Performance Metrics for Academic Abstracts (Values in %).

*6.1.2. Multi-Class LLM Detection*

To further understand the variations between LLM architectures, we conducted a multi-class detection experiment where the classifier was trained to distinguish between human-written text and text generated by different LLMs (ALLaM, Jais, OpenAI, and Llama). That is, we have a multi-class classification problem with 5 classes. Table 10 presents the performance metrics across all classes.

The multi-class detection achieved notable performance across all classes with the lowest F1-score of 94%. The model maintained high accuracy in identifying human-written text (98.62% recall, 94.49% precision), suggesting that in Arabic academic writing, human content retains distinctive characteristics despite advances in LLM capabilities. OpenAI-generated content showed the highest detectability (98.22% F1-score), followed by human text (96.51%). To further explore the classes' confusion, Figure 4 presents the confusion matrix of these classes with each other, where we can observe a marginal confusion.

| Class  | Accuracy (%) | Precision (%) | Recall (%) | F1 (%) |
|--------|--------------|---------------|------------|--------|
| Human  | 99.43        | 94.49         | 98.62      | 96.51  |
| ALLaM  | 98.07        | 97.81         | 93.55      | 95.63  |
| Jais   | 98.12        | 96.01         | 95.78      | 95.90  |
| Llama  | 97.21        | 92.48         | 95.87      | 94.14  |
| OpenAI | 99.18        | 98.57         | 97.87      | 98.22  |

**Table 10:** Multi-Class Detection Performance Metrics.

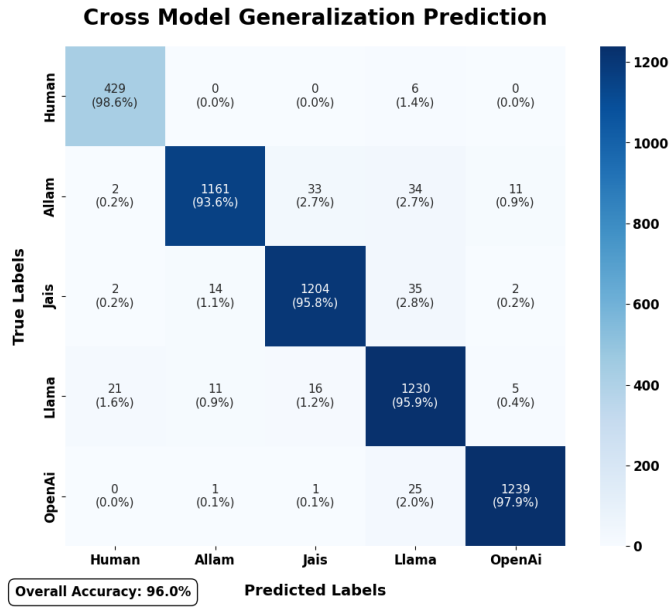

**Figure 4:** Confusion Matrix for Multi-class LLM Detection for Arabic Abstracts Dataset.

## 6.2. Social Media Content Detection

The detection experimentation on Arabic social media content presents different challenges compared to academic abstracts due to the informal na-ture (sometimes dialectical) of social media writing. This can result in more interesting detection patterns and varying performance across models. For this experiment, unlike the abstracts dataset experiments, the dataset is balanced, as we only have one generation type.

*6.2.1. Cross-Model Generalization*

In contrast to the strong cross-model performance observed with academic abstracts, the social media experiments showed relatively lower generalization capabilities, particularly in cross-architecture scenarios. Table 11 presents these results across multiple evaluation metrics. Although models, except ALLaM, show strong performance when detecting their own generations (highlighted in gray), the F1 scores (ranging from 90.69% to 98.66%) demonstrate more variation and lower values compared to academic abstracts. This suggests that even detecting a model's own generation is more challenging in this context. On the other hand, these detectors, similar to the academic abstracts experiment, maintain high precision scores. For cross-architecture detection, we can observe a notable degradation (the lowest values are highlighted in yellow). The Llama-trained detector's F1-score drops to 29.25% when testing on OpenAI generations, a much steeper decline than the academic abstracts, where cross-architecture F1-scores rarely fell below 90%. Interestingly, the OpenAI-trained detector also performed poorly when tested on Llama text, exhibiting a bidirectional performance drop. These results suggest that machine-generated text detectors may employ ensemble approaches incorporating multiple model-specific detectors for performance-critical, cross-model, reliable detection tasks.

*6.2.2. Multi-Class LLM Detection*

The multi-class detection experiment for social media content revealed variations in detectability across different LLM architectures. Table 12 presents the performance metrics for each class. From this table, we can notice that the model maintained good performance in identifying human-written text, yet with values lower than those achieved in the academic abstracts experiment. Llama demonstrates exceptionally high detectability (95.37% F1-score). Jais shows strong precision (94.21%) but lower recall (80.73%). OpenAI shows moderate performance (81.50% F1-score), and ALLaM exhibits the lowest detection rates (66.51% F1-score). These results show that different models leave distinct but varying fingerprints in their generated social media text. Llama's content stands out as the most distinctively detectable, while ALLaM's output proves notably more challenging to identify.

| Metric | Train → Test | ALLaM | Jais | Llama | OpenAI |
|---|---|---|---|---|---|
| Accuracy | ALLaM | **91.97** | 95.78 | 76.20 | 97.29 |
|  | Jais | 88.35 | **97.49** | 75.20 | 94.58 |
|  | Llama | 65.16 | 63.55 | **99.10** | 60.04 |
|  | OpenAI | 89.26 | 94.38 | 71.29 | **98.09** |
| Precision | ALLaM | **95.29** | 95.49 | 93.13 | 95.57 |
|  | Jais | 96.61 | **97.07** | 94.80 | 96.95 |
|  | Llama | 95.74 | 95.93 | **98.81** | 88.62 |
|  | OpenAI | 98.25 | 98.37 | 97.30 | **98.46** |
| Recall | ALLaM | **87.08** | 94.93 | 54.85 | 98.21 |
|  | Jais | 78.08 | **96.67** | 51.07 | 90.91 |
|  | Llama | 28.68 | 25.30 | **98.56** | 18.07 |
|  | OpenAI | 78.37 | 88.90 | 41.54 | **96.69** |
| F1-Score | ALLaM | **90.69** | 95.06 | 68.31 | 96.80 |
|  | Jais | 85.85 | **96.80** | 65.58 | 93.63 |
|  | Llama | 43.15 | 39.09 | **98.66** | 29.25 |
|  | OpenAI | 86.59 | 93.24 | 57.20 | **97.51** |

**Table 11:** Cross-Model Detection Performance for Social Media Content (Values in %).

| Class | Accuracy (%) | Precision (%) | Recall (%) | F1 (%) |
|---|---|---|---|---|
| Human | 94.38 | 80.30 | 95.85 | 87.39 |
| ALLaM | 88.23 | 71.32 | 62.31 | 66.51 |
| Jais | 94.90 | 94.21 | 80.73 | 86.95 |
| Llama | 98.07 | 95.55 | 95.18 | 95.37 |
| OpenAI | 92.69 | 78.47 | 84.78 | 81.50 |

**Table 12:** Classification Metrics for Each Class in Social Media Detection.

The confusion matrix plotted in Figure 5 shows more confusion patterns across models. We can observe that ALLaM is confused with OpenAI, jais and human text. Similarly, for Jais and OpenAI.

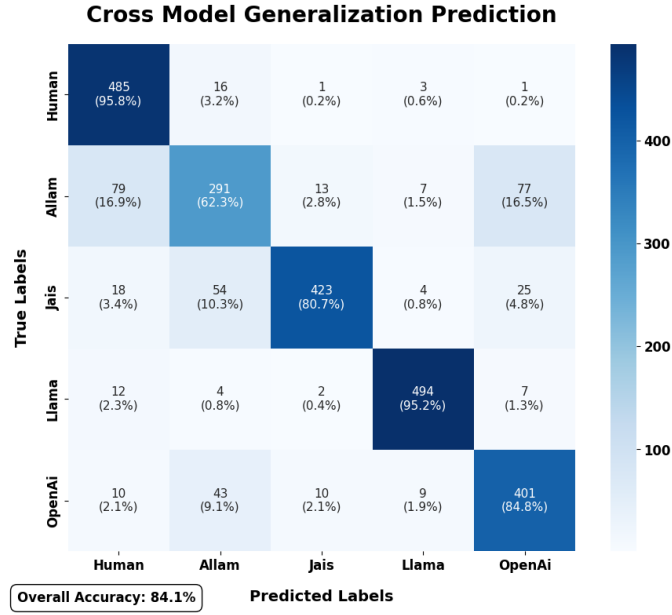

**Figure 5:** Confusion Matrix for Multi-class LLM Detection for Arabic Social Media Dataset.

## 7. Conclusion

This research investigated machine-generated Arabic text across academic and social media domains, revealing distinctive linguistic signatures that enable effective detection. Our stylometric analysis found that LLMs produce text with different vocabulary diversity and distribution. In the academic contexts, models showed steeper drop-offs in low-frequency words and higher frequency in the high-frequency words, compared to human writing. In social media contexts, LLMs consistently generated substantially shorter content (25-72% of human length) while maintaining semantic meaning, showing closer trends in the log-log frequency distribution with a notable shift compared to human writing. Our detection systems achieved excellent performance in formal contexts (99.69% F1-score for Jais, 99.9% for GPT-4) with strong cross-model generalization in academic writing, but observed degradation in social media contexts. Multi-class detection experiments further demonstrated the ability to identify specific LLMs, with varying model-specific detectability across domains. Notable challenges are still open in detecting machine-generated content in casual writing contexts, as it shows closer adherence to human text. Future work should focus on developing more robust cross-domain detection and cross-prompt methods and adaptive systems that can continuously update to address new generation methods as language models evolve.

**Acknowledgments**

We extend our deepest gratitude to the Saudi Data and AI Authority (SDAIA) and King Fahd University of Petroleum and Minerals (KFUPM) for their support through the SDAIA-KFUPM Joint Research Center for Artificial Intelligence Grant JRC-AI-RFP-20. This work would not have been possible without their substantial commitment to advancing artificial intelligence research in the Kingdom of Saudi Arabia.